%% file: main.tex
\documentclass[10pt,twocolumn,letterpaper,table]{article}

\usepackage[pagenumbers]{iccv} 
\usepackage{xcolor}
\usepackage{graphicx}
\usepackage{amsmath}
\usepackage{float}
\usepackage{amssymb}
\usepackage{booktabs}
\usepackage{adjustbox}
\usepackage{tabularx}
\newcolumntype{Y}{>{\centering\arraybackslash}X}
\usepackage{multirow}
\usepackage{xspace}
\usepackage[detect-all,per-mode=repeated-symbol]{siunitx}
\DeclareSIUnit\pixel{px}
\usepackage{microtype}
\usepackage{algorithm2e}
\usepackage{comment}
\usepackage{makecell}

\newcommand{\x}{\boldsymbol{x}}
\newcommand{\z}{\boldsymbol{z}}

\definecolor{iccvblue}{rgb}{0.21,0.49,0.74}
\usepackage[pagebackref,breaklinks,colorlinks,allcolors=iccvblue]{hyperref}

\usepackage[capitalize]{cleveref}
\crefname{section}{Sec.}{Secs.}
\Crefname{section}{Section}{Sections}
\Crefname{table}{Table}{Tables}
\crefname{table}{Tab.}{Tabs.}


\let\svthefootnote\thefootnote
\begin{document}

\title{Parameter-Efficient Adaptation of \\ Geospatial Foundation Models through Embedding Deflection} 

\author{Romain Thoreau$^\ast$\\
CNES\\
{\tt\small romain.thoreau@cnes.fr}
\and
Valerio Marsocci$^\ast$\\
European Space Agency $\Phi$-Lab\\
{\tt\small valerio.marsocci@esa.int}
\and 
Dawa Derksen \\
CNES\\
{\tt\small dawa.derksen@cnes.fr}
}

\maketitle

\begin{abstract}
As large-scale heterogeneous data sets become increasingly available, adapting foundation models at low cost has become a key issue.
Seminal works in natural language processing, \textit{e.g.} Low-Rank Adaptation (LoRA), leverage the low ''intrinsic rank'' of parameter updates during adaptation.
In this paper, we argue that incorporating stronger inductive biases in both data and models can enhance the adaptation of Geospatial Foundation Models (GFMs), pretrained on RGB satellite images, to other types of optical satellite data.
Specifically, the pretrained parameters of GFMs serve as a strong prior for the spatial structure of multispectral images.
For this reason, we introduce DEFLECT (Deflecting Embeddings for Finetuning Latent representations for Earth and Climate Tasks), a novel strategy for adapting GFMs to multispectral satellite imagery with very few additional parameters.
DEFLECT improves the representation capabilities of the extracted features, particularly enhancing spectral information, which is essential for geoscience and environmental-related tasks.
We demonstrate the effectiveness of our method across three different GFMs and five diverse datasets, ranging from forest monitoring to marine environment segmentation. Compared to competing methods, DEFLECT achieves on-par or higher accuracy with 5-10$\times$ fewer parameters for classification and segmentation tasks. The code will be made publicly available.
\end{abstract}

\let\thefootnote\relax\footnote{$^\ast$ Equal contribution.}

\addtocounter{footnote}{-1}\let\thefootnote\svthefootnote

\section{Introduction}
\label{sec:intro}
\thispagestyle{firstpage}

\begin{figure}[ht]
\centering
\includegraphics[width=\linewidth]{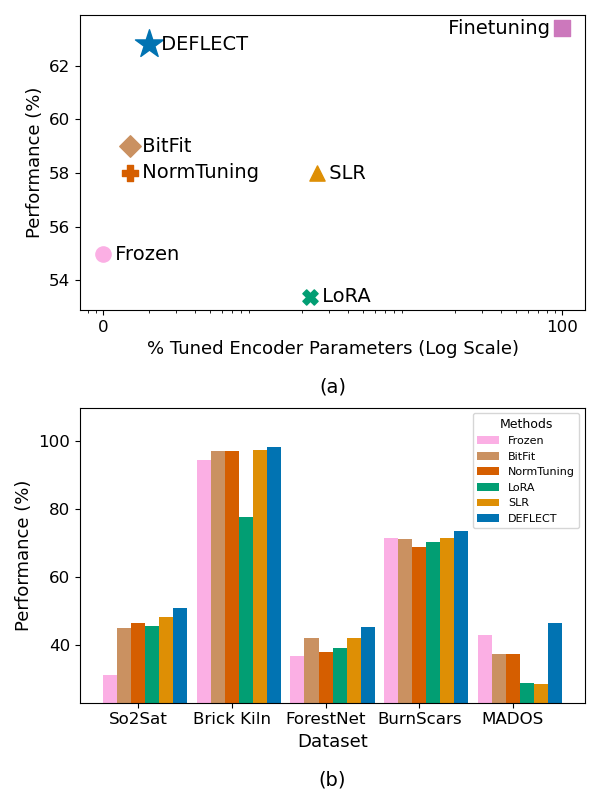}
\caption{(a) Performance (IoU and F1-score) averaged across tasks, datasets and models, against the proportion of tuned GFM parameters. Our method, DEFLECT, performs almost as well as full fine tuning (oracle), while adjusting $<1\%$ of the encoder parameters. 
(b) Performance averaged across models for each dataset.
DEFLECT consistently performs competitively for classification and semantic segmentation across five datasets, spanning from the urban domain to the forestry domain.
}
\label{fig:teaser}
\end{figure}

In recent years, the diversity and availability of satellite data have skyrocketed. 
This unprecedented access to multi/hyper-spectral satellite imagery has opened up new prospects for monitoring the Earth's biosphere. 
Real-world applications of Remote Sensing (RS) include several tasks such as agricultural monitoring from multispectral images \cite{garnot2021panoptic}, canopy traits mapping from hyperspectral images \cite{miraglio2023mapping}, and building damage assessment from very high-resolution optical imagery \cite{gupta2019xbd}.
Furthermore, combining multimodal satellite data with heterogeneous spatial, spectral and temporal resolutions has proved crucial, for instance for the detection of pluvial flood-induced damages \cite{cerbelaud2023mapping}, or for the quantification of toxic plumes released by industries and power plants \cite{calassou2024quantifying}. \vspace{0.1cm}

\noindent
Given the scarcity of ground truth on the one hand, and the massive amount of unlabeled data on the other, Geospatial Foundation Models (GFMs)\footnote{for GFM we follow the definition given in \cite{marsocci2024pangaea}} have gained considerable traction to tackle a wide variety of RS downstream tasks \cite{lacoste2021foundation, tuia2023artificial, marsocci2024pangaea}. 
GFMs are large neural networks pretrained on large-scale data sets, usually through self-supervised pretext tasks \cite{liu2024remoteclip, mendieta2023geospatial}, to learn task-generic latent representations. 
Then, task-specific predictors can be trained on top of the (frozen) features extracted by GFMs. 
However, many GFMs are pretrained on a single modality, \textit{e.g.} (very high-resolution) RGB images \cite{reed2023scalemae, liu2024remoteclip, mendieta2023geospatial}. Therefore, such models require end-to-end finetuning to handle new heterogeneous satellite data. \vspace{0.1cm}

\noindent
Conventional finetuning strategies, \textit{e.g.} end-to-end full finetuning, are resource-intensive and require the storage of large parameter sets for each task \cite{lialin2023scaling}. 
In natural language processing (NLP), Low-Rank Adaptation (LoRA) \cite{hu2021loralowrankadaptationlarge} has been a pioneering work that mitigates this issue. LoRA leverages the low "intrinsic rank" of parameter updates during adaptation to finetune large models at the cost of very few additional parameters. Many Parameter-Efficient FineTuning (PEFT) techniques in Computer Vision (CV) take inspiration from this low-rank bias \cite{zhu2024melo, fang2025dropout}. 
In RS, \cite{Scheibenreif_2024_CVPR} introduced a data-agnostic framework for adapting GFMs to new modalities via a LoRA-like technique. \vspace{0.1cm}

\noindent
In this paper, we investigate a novel paradigm for adapting GFMs in a parameter-efficient way. 
We argue that stronger inductive biases on the data, \textit{i.e.} satellite images, and models, \textit{i.e.} Vision Transformers (ViT) (that are ubiquitous in GFMs \cite{marsocci2024pangaea}), can significantly enhance the adaptation of GFMs pretrained on RGB satellite images to multispectral satellite data.
We introduce an untangled patch embedding (UPE) layer that decomposes the data embeddings into RGB embeddings and spectral embeddings.
UPE disentangles the geometric information from the radiometric information provided by multispectral channels.
Based on this decomposition, we propose a modified version of the self-attention module in the GFM, named untangled Attention (uAtt). 
Our untangled attention module allows, on the one hand, the use of the GFM pretrained parameters to process the RGB geometric and radiometric information without any updates, and on the other hand, to leverage the new radiometric information.
In order to ensure minimal disruption to the structure of the GFM's latent space, that captured high-level spatial patterns through pretraining, we introduce a principle of embedding deflection. 
Inspired by the work of \cite{karkar2021principle}, that links neural networks to dynamical systems, embedding deflection consists in deviating the trajectory of patch embeddings in the latent space of ViTs.
Our method, DEFLECT (Deflecting Embeddings for Finetuning Latent representations for Earth and Climate Tasks), requires fewer parameters than low-rank adaptation techniques. Moreover, DEFLECT can work in combination with LoRA techniques, differently from other methods.
Finally, we empirically study the benefits of DEFLECT for classification and semantic segmentation against existing adaptation methods.
The contributions of this paper are summarized as follows:
\begin{itemize}
    \item We provide \textbf{strong inductive biases} on the data and the models, tailored to multispectral satellite imagery and vision transformers.
    \item Concretely, these inductive biases translate into two new architectural elements. Firstly, a \textbf{patch embedding layer}, called UPE, which disentangles the different kinds of information of multispectral images. Secondly, \textbf{a novel attention mechanism}, called uAtt, which unties pretrained parameters and new multispectral information.
    \item We introduce a principle of \textbf{embedding deflection} that deviates the latent trajectory of the embeddings, based on the additional spectral information, while the global structure of the latent space is preserved. 
    \item We empirically demonstrate the adaptation capabilities of DEFLECT across \textbf{diverse tasks, datasets and models, with fewer parameters} than low-rank adaptation techniques.
\end{itemize}

\section{Related Works}
\label{sec:rel}

Section \ref{sec:gfm} provides an overview of state-of-the-art algorithms to pretrain GFMs, both from scratch and in a continual fashion. Section \ref{sec:peft} reviews the PEFT methods to adapt GFMs to downstream tasks and to new data domains. 

\subsection{Geospatial Foundation Models \label{sec:gfm}}

In regards to the large amount of satellite data, and to the scarcity of labels for Earth observation (EO) tasks (\textit{e.g.} land cover annotations, canopy height measurements), self-supervised learning (SSL) has become central to train GFMs \cite{wang2022self, wang2022empirical}. \vspace{0.1cm}

\noindent
\textbf{Self-supervised learning for multi-source satellite images}
Most learning algorithms rely on Masked Image Modeling (MIM) \cite{xiong2024all, cha2023billion, muhtar2023cmid, cong2022satmae}, and on Contrastive Learning (CL) \cite{wang2022dinomm, fuller2022transfer, manas2021seasonal} to learn task-agnostic latent representations. 
To cope with the heterogeneity of spatial resolutions across RGB imagery, Scale-MAE \cite{reed2023scalemae} introduced resolution-aware positional encodings, while Cross-scale MAE \cite{tang2024cross} combined contrastive and generative losses.
Other models, such as OmniSAT \cite{astruc2024omnisat} and CROMA \cite{fuller2022transfer}, combine MIM and CL to learn multimodal representations.  \vspace{0.1cm}

\noindent
\textbf{Continual pretraining} Continual pretraining involves adding new pretraining stages to an existing model to enhance or specialize its representations for specific tasks or domains. 
Initially shaped for NLP \cite{liu2021continual, gururangan2020don}, continual pretraining boosts the performance of SSL \cite{Reed_2022_WACV}. In EO, where vast unlabeled datasets are common, continual pretraining helps tailor large models for EO-specific tasks \cite{mendieta2023gfm, marsocci2024crosssensor, wang2024mtp}. Techniques like those in \cite{marsocci2023continual, moieez2023, zhang2022consecutive} combine SSL and continual learning to manage the non-stationarity of remote sensing data.
Unlike PEFT, which minimizes parameter updates to adapt models to new tasks, continual pretraining emphasizes expanding the model's core representations through ongoing updates. \vspace{0.1cm}

\noindent
Despite recent efforts, through SSL and continual pretraining, to train task-agnostic and sensor-agnostic GFMs, their generalization remains limited \cite{xu2024specialized}. 
Many recent GFMs are still confined to the processing of RGB satellite images \cite{reed2023scalemae, tang2024cross, mendieta2023gfm, liu2024remoteclip}.
Therefore, PEFT techniques are required to extend the use of GFMs to multispectral satellite images, and to a wider range of tasks, especially for environmental applications. 

\subsection{Parameter-Efficient Finetuning \label{sec:peft}}

PEFT was first introduced in NLP, aiming to adapt large pretrained models to specific tasks with minimal adjustments to the model parameters \cite{han2024parameter}. 
In RS, PEFT is of great interest as far as many specialized models need to be trained, stored, and deployed.
PEFT techniques can be divided into three main categories \cite{lialin2023scalingscaleupguide}: additive, selective, and low-rank adaptation techniques. \vspace{0.1cm}

\noindent
\textbf{Additive techniques} In NLP, \cite{houlsby2019parameter} introduced adapter modules, paving the way to additive techniques. Adapter modules extend the pretrained model with few task-specific parameters. In order to decrease the computation overhead of finetuning, \cite{pmlr-v139-fu21a}  introduced additive modules with computation and parameter sharing across multiple tasks. In RS, many PEFT techniques rely on adapter modules in order to cope with multi-source data of varying dimensions \cite{dong2024upetu, chen2023time, peft_it2023, he2024fm, airs2024}. Among them, \cite{he2024fm} add additional modules for extracting multimodal-specific representations for scene classification. \vspace{0.1cm}

\noindent
\textbf{Selective techniques} In contrast to additive methods, selective techniques finetune only a specific subset of the model parameters \cite{gheini2021cross}. Data-agnostic methods include norm-tuning \cite{zhao2023tuninglayernormattentionefficient} and BitFit \cite{zaken2021bitfit}, that only tune the layer normalization parameters and the biases, respectively, with very convincing results. Other approaches leverage sparse parameter updates, such as \cite{ansell-etal-2022-composable} that ignores the structure of the model or \cite{sung2021training} that estimates parameters' importance from the Fisher information. \vspace{0.1cm}

\noindent
\textbf{Low-Rank Adaptation} The paradigm of Low-Rank Adaptation was introduced in \cite{hu2021loralowrankadaptationlarge}. While it could be casted as an additive technique, LoRA has opened the path towards a myriad of methods \cite{malladi2023kernel, mahabadi2021compacterefficientlowrankhypercomplex, hayou2024loraefficientlowrank, ding2023sparselowrankadaptationpretrained, liu2024doraweightdecomposedlowrankadaptation} that leverage the low "intrinsic" rank of parameter updates in order to dramatically reduce the cost of finetuning. Recently, SLR \cite{Scheibenreif_2024_CVPR} proposed a combination of LoRA and a scaling strategy to adapt GFMs to new RS modalities, both in the finetuning and continual pretraining stages. \vspace{0.1cm}

\noindent
\textbf{Main limitations} On the one hand, some methods are either constrained by the tasks they address (\textit{e.g.} change detection \cite{chen2023time, li2024new, peft_it2023}, classification \cite{he2024fm}) or by the architectures they employ (e.g. the Segment Anything Model \cite{kirillov2023segment} in \cite{mesam2024, chen2023time}). 
On the other hand, many techniques inspired by research in NLP do not address the specificities of CV or remote sensing.
Our strategy, DEFLECT, leverages inductive biases on the data and on the ubiquitous ViT, such that it can be applied to any ViT-based GFM and to any downstream task. 
\section{Preliminary \& Motivation}

\noindent
\textbf{Setup} 
Let us consider a downstream task, for which we have a labeled dataset of multispectral images. In order to solve the task, consider using a ViT-based GFM, pretrained on RGB satellite images, which typically have higher spatial resolution than multispectral images. Finally, consider one instance from the dataset, \textit{i.e.} a satellite multispectral image $\boldsymbol{X} \in \mathbb{R}^{C \times H \times W}$, that comprises $H \times W$ pixels with $C$ spectral channels.\\

\noindent
\textbf{Vision Transformers} Data processing in ViTs can be considered in three stages \cite{karkar2021principle}: i) dimensionality change, ii) data transport, and iii) task-specific predictions. First, the input image $\boldsymbol{X}$ is divided into $n$ patches. Each patch is encoded into a $d$-dimensional embedding, resulting in a latent representation  $\x = [\x_1 \: \ldots, \x_n]^T \in \mathbb{R}^{n \times d}$. Second, ViTs transport the embeddings in the latent space through $m$ attention blocks. We denote the input of the $l^{th}$ attention block as $\z^{(l)}$, \textit{e.g.} $\z^{(1)} = \x$. The $l^{th}$ layer computes the displacement from $\z^{(l)}$ to $\z^{(l+1)}$. Third, task-specific layers map the latent representations to predictions. 
Attention blocks themselves process the data in two steps. First, a self-attention module, comprising query, key and value matrices denoted as $W_l^Q$, $W_l^K$, $W_l^V \in \mathbb{R}^{d \times d}$, respectively, computes a first displacement:
\begin{align}
    \Delta_1 \z_i^{(l)} = \sum_{j=1}^n \frac{\mbox{exp}(\alpha_{ij})}{\sum_{j'=1}^n \mbox{exp}(\alpha_{ij'})}(\z_j^{(l)}W_l^V), \\
    \mbox{ where } \alpha_{ij} = \frac{1}{\sqrt{d}}(\z_i^{(l)}W_l^Q)(\z_j^{(l)}W_l^K)^T. \label{eq:attn-prod}
\end{align}
Second, a standard MLP computes a second displacement:
\begin{align}
    \Delta_2 \z_i^{(l)} = \mbox{MLP}^{(l)}(\z_i^{(l)} + \Delta_1 \z_i^{(l)}) 
\end{align}
All in all, the embeddings are transported in the latent space as follows:
\begin{align}
    \z_i^{(l+1)} = \z_i^{(l)} + \Delta_1 \z_i^{(l)} + \Delta_2 \z_i^{(l)}
\end{align}
For readibility, we omitted the normalization layers, and considered a single-head self-attention module. \vspace{0.1cm}

\noindent
\textbf{Embedding decomposition} In multispectral images, spectral channels beyond RGB provide crucial information for many environmental applications. 
The additional radiometric information is indeed highly informative of the physical and chemical properties of the Earth's biosphere.
The geometric information of the additional spectral channels, however, is redundant with the geometric information of the RGB channels.
In that regard, we assume that the patch embeddings $\x$, computed by an arbitrary neural network, can be decomposed as follows:
\begin{equation}
    \x = \x_P + \x_A \label{eq:embedding_decomposition}
\end{equation}
where $\x_P$ encodes the radiometric and geometric information of the RGB channels, while $\x_A$ encodes the spectral information beyond RGB. 
Practically, $\x_P$ is the patch embedding computed by the \textbf{p}retrained foundation model. On the contrary, $\x_A$ corresponds to \textbf{a}uxiliary information that the pretrained model is unable to handle.  \vspace{0.1cm}

\noindent
\textbf{Limitations of low-rank methods} Low-rank methods finetune a small amount of new parameters, denoted as $\Delta W$, in addition to pretrained parameters, denoted as $W_P$, resulting in task-specific parameters $\tilde{W}$:
\begin{align}
    \tilde{W} = W_P + \Delta W
\end{align}
Therefore, assuming the existence of the embedding decomposition defined in Eq. \ref{eq:embedding_decomposition}, the attention product of the first attention block, defined in Eq. \ref{eq:attn-prod}, becomes\footnote{For readability, we omitted the superscript $(1)$ indicating the first layer.}:
\begin{align}
    \alpha_{ij}^{\mbox{\tiny{LR}}} & = \frac{1}{\sqrt{d}}((\x_{iP} + \x_{iA})\tilde{W}^Q)((\x_{jP} + \x_{jA})\tilde{W}^K)^T \nonumber \\
    & = \frac{1}{\sqrt{d}}\big[(\x_{iP}\tilde{W}^Q)(\x_{jP}\tilde{W}^K)^T + (\x_{iP}\tilde{W}^Q)(\x_{jA}\tilde{W}^K)^T \nonumber \\
    & + (\x_{iA}\tilde{W}^Q)(\x_{jP}\tilde{W}^K)^T + (\x_{iA}\tilde{W}^Q)(\x_{jA}\tilde{W}^K)^T\big] \label{eq:lim_1}
\end{align}
Eq. \ref{eq:lim_1} shows that the patch embeddings $\x_P$ and $\x_A$ are entangled. Therefore, the projection matrices $\tilde{W}^Q$ and $\tilde{W}^K$ are shared across embeddings. However, $\x_P$ and $\x_A$ encode two kinds of information which are heterogeneous. Therefore,  applying the same projection does not seem reasonable. To be precise, let us expand the first and last terms of Eq. \ref{eq:lim_1}:
\begin{align}
    \alpha_{ij}^{\mbox{\tiny{LR}}} & = \frac{1}{\sqrt{d}}\big[(\x_{iP}W_P^Q)(\x_{jP}W_P^K)^T + (\x_{iP}W_P^Q)(\x_{jP}\Delta W^K)^T \nonumber \\
    & + (\x_{iP}\Delta W^Q)(\x_{jP}W_P^K)^T + (\x_{iP}\Delta W^Q)(\x_{jP}\Delta W^K)^T \nonumber \\
    & + \ldots + (\x_{iA}W_P^Q)(\x_{jA}W_P^K)^T + (\x_{iA}W_P^Q)(\x_{jA}\Delta W^K)^T \nonumber \\
    & + (\x_{iA}\Delta W^Q)(\x_{jA}W_P^K)^T + (\x_{iA}\Delta W^Q)(\x_{jA}\Delta W^K)^T \big]    \label{eq:attn-expansion}
\end{align}
Eq. \ref{eq:attn-expansion} shows that low-rank methods use pretrained parameters $W_P$ to indifferently process the embeddings $\x_P$ and $\x_A$, although $\x_A$ encodes spectral information that was not available during pretraining. Similarly, the new parameters $\Delta W$ handle both $\x_P$ and $\x_A$, whereas we would expect the new parameters to be dedicated to handling $\x_A$. Thus, we argue that low-rank adaptation methods, due to the linear arithmetic of the self-attention module between $\x_P$ and $\x_A$, may not fully leverage either the additional spectral information of multispectral images or the pretrained parameters of the GFM, which encode RGB information.

\section{Methodology}
\label{sec:meth}

\begin{figure*}[t]
    \centering
    \includegraphics[width=\linewidth]{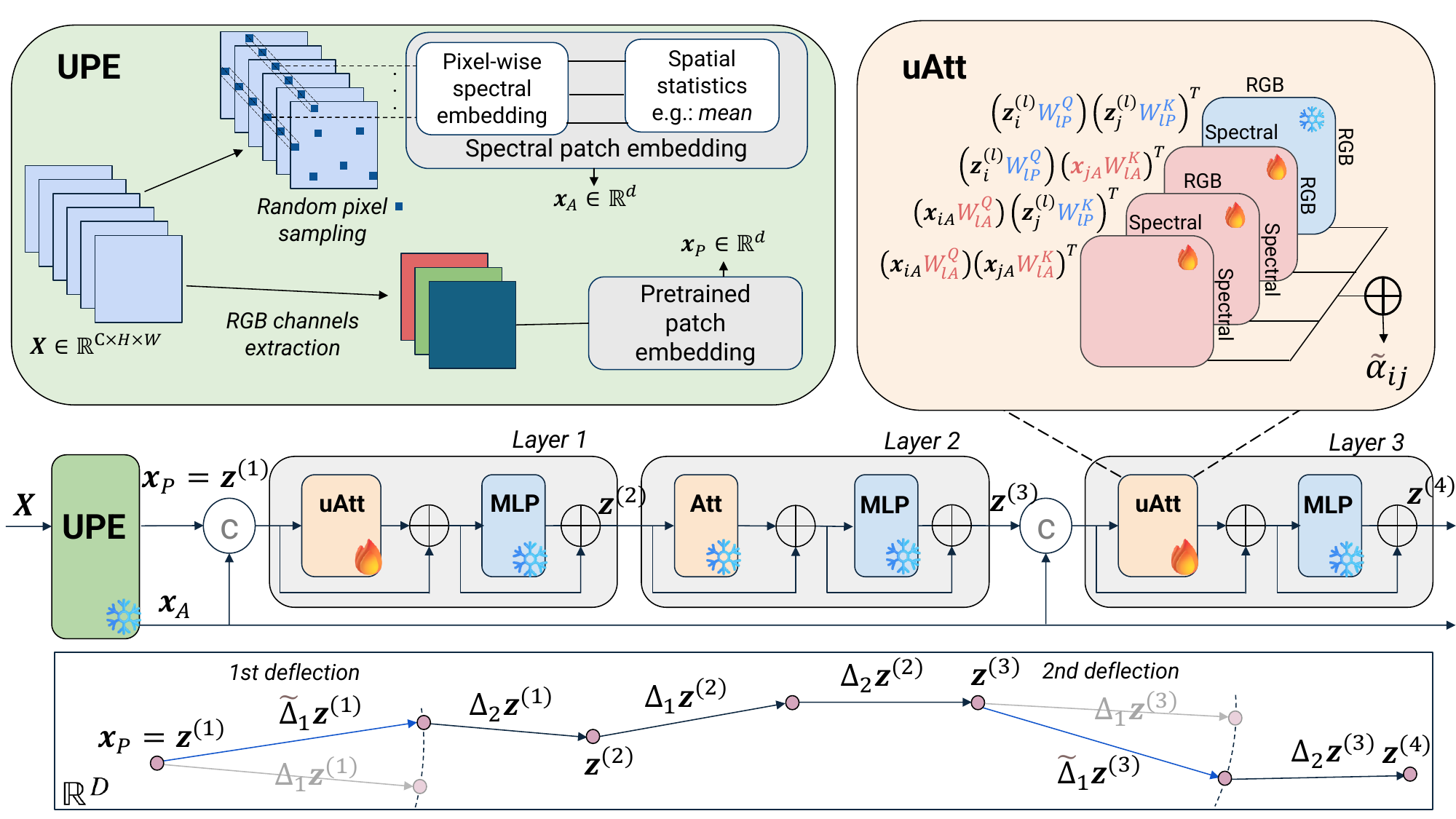}
    \caption{Illustration of DEFLECT, our parameter-efficient adaptation technique. The upper-left block describes UPE, our untangled patch embedding layer, introduced in section \ref{sec:untangled_patch}. UPE computes the RGB embeddings $\x_P$ and the auxiliary embeddings $\x_A$. The upper-right block illustrates uAtt, our untangled attention module, introduced in section \ref{sec:untangled_attention}. The middle part of the figure illustrates the first three attention layers of the GFM. In this example, first and third layers are adapted: the standard attention modules are replaced by uAtt.
    Adapted layer $l$ takes as input the concatenation of $\z^{(l)}$ and $\x_A$.
    The additional radiometric information encoded into $\x_A$ is leveraged by uAtt in order to deflect the trajectory of the embeddings. The displacement $\Delta_1 \z^{(l)}$ that the attention block with pretrained parameters would have computed is deflected, resulting in a new displacement $\tilde{\Delta}_1 \z^{(l)}$. The norm of $\tilde{\Delta}_1 \z^{(l)}$ is constrained to be equal to the norm of $\tilde{\Delta}_1 \z^{(l)}$, as illustrated in the bottom part of the figure.
    }
    \label{fig:deflect}
\end{figure*}

This section describes DEFLECT, our parameter-efficient adaptation strategy, to adapt GFMs, pretrained on RGB satellite images, to multispectral satellite images.
DEFLECT, illustrated in Fig. \ref{fig:deflect}, is an additive technique that complements pretrained parameters with new parameters, in order to integrate the additional spectral information of multispectral images. 
It comprises three key elements:
\begin{itemize}
    \item An \textbf{Untangled Patch Embedding Layer}, called \textbf{UPE}, [section \ref{sec:untangled_patch}] that computes the patch embeddings $\x_P$ and $\x_A$ separately,
    \item A novel \textbf{Untangled Attention Block}, called \textbf{uAtt}, [section \ref{sec:untangled_attention}] that integrates the new spectral information,
    \item An \textbf{Embedding Deflection Mechanism} [section \ref{sec:deflect}] that deflects the trajectory of patch embeddings $\x_P$, transported by pretrained attention blocks, based on the untangled attention product. 
    The embedding deflection aims to increase the representational power of the latent embeddings.
\end{itemize}

\subsection{Untangled Patch Embedding \label{sec:untangled_patch}}

Appending multispectral channels to RGB images can be thought as adding color to grayscale images.
New spectral channels provide informative radiometric information but redundant geometric information.
Therefore, we focus on extracting discriminative spectral features from new multispectral channels, and rely on the pretrained model to extract geometric features (and RGB radiometric features).
Our \textit{Untangled Patch Embedding} is, thus, made of two parts:
\begin{itemize}
    \item \textbf{Embedding of $\x_P$.} We extract RGB channels from the multispectral image and compute $\x_P$ with the pretrained patch embedding layer, from the pretrained GFM.
    \item \textbf{Embedding of $\x_A$.} As we are only interested in radiometric information, we compute $\x_A$ with a pixel-set encoding technique \cite{garnot2020satellite}. First, we randomly sample a set of pixels (10\% of pixels within a patch in our experiments). Second, we compute pixel-wise spectral indices (\textit{i.e.} \textit{normalized} linear combinations of spectral channels) such as NDVI \cite{rouse1973paper} or NDTI \cite{LACAUX200766}, in order to extract spectral features without overloading the model with new parameters.  Third, we compute and concatenate statistics of the spectral features (\textit{e.g.} mean, standard deviation, first and last quartiles, minimum and maximum...). 
    In this way, the dimension of $\x_A$ is invariant to the size of the patch and the number of sampled pixels. Moreover, we can save memory and computation, without the need of introducing another learnable patch embedder.
\end{itemize}
Our Untangled Patch Embedding is illustrated in the upper-left block of Fig. \ref{fig:deflect}.

\subsection{Untangled Attention Block \label{sec:untangled_attention}}

Inspired by \cite{kerethinking}, we introduce \textit{uAtt}, an untangled self-attention module that mitigates the effects of the linear arithmetic of self-attention.
uAtt modules partially replace standard attention modules, for a subset of layers. 
At layer $l$, uAtt leverages pretrained parameters $W_{lP}$ alongside new parameters $W_{lA}$, in order to compute a modified displacement $\tilde{\Delta}_1$, instead of $\Delta_1$ in eq. \ref{eq:attn-prod}.
Basically, parameters $W_{lP}$ and $W_{lA}$ are meant to process the untangled embeddings computed by our UPE layer:
\begin{align}
    \tilde{\Delta}_1 & \z_i^{(l)} = \sum_{j=1}^n \frac{\mbox{exp}(\tilde{\alpha}_{ij})}{\sum_{j'=1}^n \mbox{exp}(\tilde{\alpha}_{ij'})}(\z_{j}^{(l)}W_{lP}^V + \x_{jA}W_{lA}^V); \\
    \tilde{\alpha}_{ij} & = \frac{1}{\sqrt{d}} (\z_i^{(l)}W_P^Q + \x_{iA}W_A^Q)(\z_j^{(l)}W_P^K + \x_{jA}W_A^K)^T \label{eq:uatt1} 
\end{align}
At first layer, pretrained -- and not updated during finetuning -- parameters $W_{1P}$ process the embeddings $\x_P = \z^{(1)}$.
Then, at layer $l > 1$, parameters $W_{lP}$ (not updated either) process the embeddings $\z^{(l)}$.
Those operations result in usual RGB-to-RGB attention products, as in the pretrained GFM.
In addition, new parameters $W_{lA}$ process the new spectral information encoded into $\x_A$, which is kept fixed along every layers.
This yields extra RGB-to-spectral, spectral-to-RGB and spectral-to-spectral attention products:
\begin{align}
    \tilde{\alpha}_{ij} = & \frac{1}{\sqrt{d}}\big[\overbrace{(\z_{i}^{(l)}W_{lP}^Q)(\z_{j}^{(l)}W_{lP}^K)^T}^{\mbox{\scriptsize{RGB-to-RGB}}} + \overbrace{(\z_{i}^{(l)}W_{lP}^Q)(\x_{jA}W_{lA}^K)^T}^{\mbox{\scriptsize{RGB-to-spectral}}} \nonumber \\
    & + \underbrace{(\x_{iA}W_{lA}^Q)(\z_{j}^{(l)}W_{lP}^K)^T}_{\mbox{\scriptsize{spectral-to-RGB}}} + \underbrace{(\x_{iA}W_{lA}^Q)(\x_{jA}W_{lA}^K)^T}_{\mbox{\scriptsize{spectral-to-spectral}}} \big] \label{eq:uatt} 
\end{align}
uAtt is illustrated in the upper-right of Fig. \ref{fig:deflect}. The new parameters of the untangle attention module can, optionally, be re-parameterized in a low-rank fashion.

\subsection{Embedding Deflection \label{sec:deflect}}

In order to better separate ambiguous latent clusters, we aim to leverage the new radiometric information of multispectral images. 
To this end, we adapt a subset of the attention blocks, denoted as $\mathcal{K}$, leading to few additional parameters.
The attention module of adapted blocks is replaced by our uAtt module.
Furthermore, we aim to preserve the structure of the pretrained GFM latent space, and thus the capacity of the GFM to process the geometric, textural, and contextual information of multispectral images.
Therefore, we constrain the norm of the displacement computed by uAtt to match the norm of the displacement that the standard attention block would have computed.
All in all, given a labeled training data set $\mathcal{D}_T$, a classification or a semantic segmentation loss $\mathcal{L}$, and a predictor head $p_\phi$ that maps the latent embeddings to the target classes, we define the adaptation problem as follows:
\begin{align}
    \min_{\theta_A, \phi} & \:\:\:\: \mathcal{L}(\theta_A, \phi; \theta_P, \mathcal{D}_T) \label{eq:finetuning} \\
    \mbox{subject to} & \:\:\:\: \|\Delta_1 \z_i^{(k)}\|_2 = \|\tilde{\Delta}_1 \z_i^{(k)}\|_2, \mbox{for all } k \in \mathcal{K} \nonumber 
\end{align}
where $\theta_P$ is the set of all pretrained parameters, and $\theta_A$ is the set of new query, key, and value matrices.
In practice, the hard constraints in Eq. \ref{eq:finetuning} can be satisfied by tweaking the architecture of the untangle attention module, only at the cost of running the pretrained GFM on the RGB channels to compute the norm of $\Delta_1\z$. This principle of embedding deflection is illustrated in the bottom part of Fig. \ref{fig:deflect}.

\section{Experiments}

\subsection{Experimental Setup}
\label{sec:exps}

\begin{table}[t]
\caption{Overview of selected geospatial datasets used for classification and segmentation tasks.
\label{tab:data}}
\resizebox{\columnwidth}{!}{%
\begin{tabular}{llccc}
\hline
Dataset     & Task                & Bands & Patch Size & Samples \\ \hline
\multicolumn{5}{c}{Classification} \\ \hline
So2Sat      & Urban dynamics      & 10    & 32x32      & 22,000          \\
Brick Kiln  & Brick kilns         & 13    & 64x64      & 19,000          \\
ForestNet   & Forest monitoring   & 12    & 332x332    & 8,000           \\ \hline
\multicolumn{5}{c}{Segmentation} \\ \hline
MADOS       & Marine pollution    & 13    & 240x240    & 3,000           \\
BurnScars   & Burn scars          & 6     & 512x512    & 1,000           \\ \hline
\end{tabular}
}
\end{table}

\noindent 
\textbf{Datasets} To benchmark DEFLECT, we selected three classification datasets and two segmentation datasets (cf. Tab \ref{tab:data}). All the datasets consist of satellite multispectral images with target data related to environmental sciences. 
Note that for the classification datasets, we adopted the GEO-Bench \cite{lacoste2023geobench} preprocessing. \vspace{0.1cm}

\noindent 
\textbf{Pretrained GFMs} We evaluated the performance of DEFLECT on three GFMs, pretrained on RGB images, each selected for its distinct characteristics. The first model is Scale-MAE \cite{reed2023scalemae}, a variant of MAE widely used for RS representation learning. The second model, DINO-MC \cite{wanyan2024extending}, is a contrastive approach based on DINO. The third model, Cross-Scale MAE \cite{tang2024cross}, was pretrained with a hyrid objective which combines both a MAE and a contrastive approach. \vspace{0.1cm}

\noindent 
\textbf{Competing methods} We compared the performance of DEFLECT against several competing PEFT techniques. The baseline uses a frozen encoder. The oracle represents the upper bound, where the entire model is finetuned end-to-end. Additionally, we evaluated DEFLECT against four widely used, open-source, both generalist and specialized PEFT approaches: LoRA \cite{hu2021loralowrankadaptationlarge}, Scaled Low-Rank (SLR) \cite{Scheibenreif_2024_CVPR}, BitFit \cite{zaken2021bitfit}, and NormTuning \cite{zhao2023tuninglayernormattentionefficient}. \vspace{0.1cm}

\noindent
\textbf{DEFLECT implementation} For DEFLECT training, the spectral patch embeddings were computed with the following statistics: mean, standard deviation, minimum, maximum, the first and third quartiles, as well as the 0.1/0.4/0.6/0.9-quantiles . In order to align the embedding dimension of the spectral encoder with that of the RGB feature extractor, we further projected the extracted feature with a linear layer. This layer was initialized with a Xavier initialization, followed by a Layer Normalization and a GELU activation function in a Transformer fashion. We replaced the traditional attention blocks with our uAtt in four layers, the ones used by UPerNet (\textit{i.e.} 3, 5, 7, 11 for ViT Base; 7, 11, 15, 23 for ViT Large).
The additional parameters of the uAtt blocks are low-rank (with a rank of 16). \vspace{0.1cm}

\begin{table*}[t]
    \caption{Comparison of F1-score (classification) and IoU (segmentation) results across downstream tasks for DEFLECT and competing PEFT methods on GFMs with different approaches. Each method is evaluated across diverse datasets representing geospatial challenges: So2Sat, Brick Kiln, ForestNet, Burn Scars, and MADOS. Average classification performance, average segmentation performance, and overall performance are also reported. DEFLECT consistently demonstrates robust performance with substantially fewer trainable parameters across self-supervised models and tasks, achieving top scores in multiple datasets and a good balance between efficiency and performance. The Oracle finetuning performance is included as a top reference.}
    \centering
    \begin{tabular}{lccccccccc}

         Method & \makecell{Encoder\\Tuned Params} & \makecell{So2Sat \\ (mF1)} & \makecell{Brick Kiln\\ (mF1)} & \makecell{ForestNet\\ (mF1)} & \makecell{Burn Scars\\ (IoU)} & \makecell{MADOS\\ (mIoU)} & \makecell{Avg. \\Class.} & \makecell{Avg. \\Segm.} & \makecell{Avg. \\Perf.}\\ \toprule
         
         \multicolumn{9}{c}{Scale-MAE \cite{reed2023scalemae}} \\ \hline
         \rowcolor{gray!20} Finetuning (oracle) & 100\% & 52.6 & 98.7 & 45.1 & 79.1 & 47.0 & 65.4& 63.0 & 64.5\\
         Frozen & 0.0\% & 31.5 & 94.1 & 41.8& 76.2 & 36.0 & 55.8& 56.1 & 55.9 \\
         Norm Tuning \cite{zhao2023tuninglayernormattentionefficient} & 0.03\% &  51.7& 97.5 & 42.8 & 68.4 & 18.4 & 64.0 & 43.4 & 55.8\\
         BitFit \cite{zaken2021bitfit} & 0.09\% & 46.4& \underline{97.9} & 42.7 & 76.0 & 19.4 & 62.3& 47.7 & 56.5\\
         LoRA \cite{hu2021loralowrankadaptationlarge} & 2.1\% & \textbf{54.9} & 97.5 & 42.6 & \underline{79.5} & 45.0 & \textbf{65.0}& 62.2 & 63.9\\
         SLR \cite{Scheibenreif_2024_CVPR} & 2.2\% & 51.3 & \textbf{98.3} & \underline{44.0} & \textbf{80.1} & \underline{46.5} & 64.5& \underline{63.3} & \underline{64.0}\\
         \textbf{DEFLECT (ours) }& 0.2\% & \underline{53.2}& 97.8 & \textbf{44.1} & 77.3 & \textbf{50.6} & \textbf{65.0}& \textbf{64.0}& \textbf{64.6}\\ \hline
         
         \multicolumn{9}{c}{DINO-MC \cite{wanyan2024extending}} \\ \hline
         \rowcolor{gray!20} Finetuning (oracle) & 100\% & 48.8 & 98.9 & 44.2 & 76.5 & 61.6 & 64.0& 69.0 & 66.0\\ 
         Frozen & 0.0\% & 33.2 & 94.6 & 34.5 & 70.4 & 51.8 & 54.1& 61.1& 56.9\\
         Norm Tuning \cite{zhao2023tuninglayernormattentionefficient} & 0.09\% & 47.2 & 98.1 & 38.6 & 70.2 &  \textbf{53.8} & 61.3 & \underline{62.0} & {61.6}\\
         BitFit \cite{zaken2021bitfit} & 0.2\% & 49.1& 98.3 & \textbf{46.1} & 69.8 & \underline{53.0} & \underline{64.5}& 61.4 & \underline{63.3}\\
         LoRA \cite{hu2021loralowrankadaptationlarge} & 5.4\% & 47.1 & 39.9 & 40.2 & 65.5 & 6.3 & 42.4& 35.9 & 39.8\\
         SLR \cite{Scheibenreif_2024_CVPR} & 5.7\% & \underline{48.8} & \underline{98.3} & {42.4} & \underline{70.5} & 3.5 & {63.2} & 37.0 & {52.7}\\
         \textbf{DEFLECT (ours)}& 0.9\% & \textbf{52.4} & \textbf{98.8} & \underline{45.4} & \textbf{75.6} & 51.6 & \textbf{65.5}&  \textbf{63.6} & \textbf{64.8}\\ \hline
         
         \multicolumn{9}{c}{Cross-Scale MAE \cite{tang2024cross}} \\ \hline
         \rowcolor{gray!20} Finetuning (oracle) & 100\% & 45.1& 98.3& 37.1 & 78.1 & 47.2 & 60.2& 62.7 & 61.2\\
         Frozen & 0.0\% & 29.9& 94.8& 33.5 & 68.0 & \textbf{41.4} & 52.7& \textbf{54.7} & 53.5\\
         Norm Tuning \cite{zhao2023tuninglayernormattentionefficient} & 0.03\% & 41.2 & 96.8 & 32.4 & \underline{68.4} & \underline{40.7} & 56.8 & \underline{54.6} & 55.9\\
         BitFit \cite{zaken2021bitfit} & 0.09\% & 40.2& 95.6 & 38.8 & 67.9 & 40.3 & 58.2& 54.1 & \underline{56.6}\\
         LoRA \cite{hu2021loralowrankadaptationlarge} & 2.1\% & 33.0 & 96.1& 34.5 & 66.4 & 35.0 & 54.6& 50.7 & 53.0\\
         SLR \cite{Scheibenreif_2024_CVPR} & 2.2\% & \underline{44.4}& \underline{96.2}& \underline{41.8} & 63.7 & 35.6 & \underline{60.8}& 49.7 & 56.3\\
         \textbf{DEFLECT (ours)}& 0.2\% & \textbf{45.9}& \textbf{98.4}& \textbf{46.3} & \textbf{70.6} & 38.2 & \textbf{63.5}& 54.4 & \textbf{59.9}\\
    \end{tabular}
    \label{tab:metrics}
\end{table*}

\noindent
\textbf{Evaluation protocol} For the classification tasks, we used a linear layer and reported the mean F1-score. Models are trained for 60 epochs with a batch size of 24. For the segmentation tasks, we used a trainable UPerNet as the decoder and reported Intersection over Union (IoU) for 2-classes datasets and mean IoU (mIoU) otherwise. The segmentation models are finetuned for 120 epochs with a batch size of 8. In all experiments, all the images are 224x224 pixels to match the encoder input size (they are either upsampled or random cropped). The learning rate is set to 0.0001 and optimized using the AdamW optimizer, with a MultiStep scheduler that applies learning rate adjustments at steps 0.6 and 0.9 of the total number of epochs. The loss function used for both classification and segmentation tasks is the cross entropy loss. 
The code will be made available for reproducibility purposes.

\subsection{Experimental Results}
\label{sec:expr}

Table \ref{tab:metrics} presents the classification and segmentation performance of DEFLECT and competing methods. We report several key observations in the following. \vspace{0.1cm}

\noindent
\textbf{Stability across tasks, datasets and models} We observe that every competing methods have an  erratic performance across the tasks, datasets and models. For instance, NormTuning reached on MADOS the best mIoU with DINO-MC (53.8 mIoU / +2.2 pt compared to DEFLECT) while it achieved the worst with Scale-MAE (18.4 mIoU / -32.2 pt compared to DEFLECT). LoRA and SLR have also exhibited very large performance gaps, for instance on MADOS as well, on which their mIoU collapsed with DINO-MC. In contrast, DEFLECT demonstrated stable results across tasks, datasets and models, reaching the best performance in average (e.g. 64.0 mIoU with Scale-MAE), as well as for many configurations. In the Supplementary Material, we empirically study the link between performance consistency and embedding displacement.\vspace{0.1cm}

\noindent
\textbf{Parameter efficiency} DEFLECT showcased a very good trade-off between the number of updated parameters and performance. As a matter of fact, DEFLECT is often in par, and sometimes better, than full finetuning (see Figure \ref{fig:teaser}). For instance, DEFLECT outperformed full finetuning with Scale-MAE on MADOS by 3.6 pt and with Cross-Scale MAE on ForestNet by 9.2 pt. While the low-rank adaptation techniques, LoRA and SLR, have 2.1\% tunable parameters for Scale-MAE, DEFLECT only tunes 0.2\% of the model parameters for comparable or better performance. \vspace{0.1cm}

\noindent
Overall, the performance of DEFLECT alongside its stability and parameter-efficiency support the potential of our embedding decomposition and embedding deflection technique.

\subsection{Ablation studies}

In the following, we conduct ablation studies about i) the best time to deflect the latent trajectories, and ii) about performance across different low-rank dimensions. In the Supplementary Material, we report additional ablations about the number of pixels sampled in UPE, about reprojecting the features computed by UPE with a linear layer, and about the inconsistency of the parameters initialization strategies for competitive methods w.r.t DEFLECT. In addition, the Supplementary Material gives more details about the datasets and presents extensive results. \vspace{0.1cm}

\noindent
\textbf{Best time for embedding deflection?} We argue that the embeddings can be deflected only a few times. In other words, only a few layers of the GFM have to be adapted. Experimental results, presented in Tab. \ref{tab:adapted_times}, actually show that deviating embeddings at each time step (\textit{i.e.} each layer), decreases the performance. On the contrary, only adapting the first layer leads to suboptimal performance. We argue that deflecting the embeddings at the first layer only is deflecting too little, not including enough spectral information in the embeddings, while deflecting the embeddings on every layer changes too much the trajectory of the embeddings. \vspace{0.1cm}

\begin{table}[ht]
\caption{Ablation study about the GFM layers to be adapted, for each model on MADOS / BurnScars. 
\label{tab:adapted_times}}
\centering
\resizebox{\columnwidth}{!}{
\begin{tabular}{l c c c}
& \multicolumn{3}{c}{Adapted Layers} \\ \cline{2-4}
Model & First Layer & UPerNet Layers & All layers \\ \toprule
Scale-MAE & 47.1 / 75.8 & \textbf{50.6} / \textbf{77.3} & 38.7 / 70.1 \\
DINO-MC & 50.8 / {73.7} & \textbf{51.6} / \textbf{75.6} & 47.1 / 67.8 \\
Cross-Scale MAE & \textbf{39.2} / {67.4} & 38.2 / \textbf{70.6} & 31.7 / 64.5 \\
\end{tabular}
}
\end{table}

\noindent
\textbf{Performance across low-rank dimensions} Table \ref{tab:lr} compares DEFLECT and LoRA across different models and low-rank dimensions. The results show that DEFLECT achieves competitive or superior performance while tuning significantly fewer parameters. For instance, in Scale-MAE with LR dim = 32, DEFLECT attains 51.3 mIoU (MADOS) and 77.3 IoU (BurnScars) while tuning only 0.28\% of parameters, compared to 30.2 / 78.7 with LoRA, which requires 4.1\%. Similar trends are observed for DINO-MC and Cross-Scale MAE.
Besides, DEFLECT without low-rank adaptation is better than with low-rank adaptation. It confirms that DEFLECT is inherently independent of low-rank techniques and does not rely on low-rank adaptation to be effective.

\begin{table}[]
\caption{Performance comparison of DEFLECT / LoRA for different models (\#1 Scale-MAE,\#2 DINO-MC, \#3 Cross-Scale MAE) across varying low-rank dimensions. The results indicate minimal differences for MADOS (mIoU) and BurnScars (IoU), highlighting the potential for selecting a low-rank dimension that balances computational efficiency and performance.  The superscript $^\ast$ indicates the default dimension used for the main experiments.}
\label{tab:lr}

\begin{tabular}{ccccc}
& & \multicolumn{3}{c}{DEFLECT / LoRA} \\ \cmidrule{3-5}
Model &
  \begin{tabular}[c]{@{}c@{}}LR\\ dim\end{tabular} &
  \begin{tabular}[c]{@{}c@{}} Tuned \\ params (\%)\end{tabular} &
  \begin{tabular}[c]{@{}c@{}}MADOS \\ (mIoU)\end{tabular} &
  \begin{tabular}[c]{@{}c@{}}BurnScars\\ (IoU)\end{tabular} \\ \midrule
\multirow{4}{*}{\#1}            & 8    & \textbf{0.15} / 1.1 &    \textbf{50.7} / 47.0  &   77.4 /  \textbf{79.3}  \\ 
                                & 16$^\ast$  &  \textbf{0.20} / 2.1 &  \textbf{50.6} / 45.0  & 77.3 /  \textbf{79.5} \\
                                & 32   &  \textbf{0.28} / 4.1 &  \textbf{51.3} / 30.2 & 77.3 /  \textbf{78.7} \\
                                & None & 4.1 / N.A. & 51.4 / N.A. & 77.9 / N.A. \\ \midrule
\multirow{4}{*}{\#2}            & 8    &  \textbf{0.51} / 2.9 &  \textbf{50.9} / 4.0 &  \textbf{75.5} / 72.7 \\ 
                                & 16$^\ast$ &  \textbf{0.9} / 5.4 &  \textbf{51.6} / 6.3 &  \textbf{75.6} / 65.5 \\
                                & 32 &  \textbf{1.2} / 10.0 &  \textbf{52.3} / 7.6 &  \textbf{74.3} / 26.4  \\
                                & None & 7.8 / N.A. & 52.9 / N.A.  & 76.1 / N.A. \\ \midrule
\multirow{4}{*}{\#3}            & 8 &  \textbf{0.15} / 1.12 &  37.1 /  \textbf{39.5} & 70.0 /  \textbf{74.3}  \\ 
                                & 16$^\ast$ &  \textbf{0.20} / 2.1 &  \textbf{38.2} / 35.0 &  \textbf{70.6} / 66.4 \\
                                & 32 &  \textbf{0.28} / 4.1 &  \textbf{37.9} / 26.1 &  \textbf{71.8} / 60.2 \\
                                & None & 4.1 / N.A. & 38.6 / N.A. & 72.1 / N.A. \\ 
\end{tabular}
\end{table}

\section{Conclusions and perspectives}
\label{sec:conc}

In this work, we introduced DEFLECT, a novel approach for adapting Geospatial Foundation Models to multispectral satellite images. 
In contrast to Low-Rank Adaptation techniques such as SLR \cite{Scheibenreif_2024_CVPR}, we leveraged inductive biases on the data and on the models, introducing an untangled patch embedding layer and an untangled attention module. 
Through a principle of embedding deflection, DEFLECT integrates the additional radiometric information of multispectral data, at the cost of very few parameters, in order to increase the expressiveness of the GFM latent space.
Experiments across several tasks, datasets and models, demonstrated that DEFLECT reached high and consistent accuracy, compared with competing methods that performed erratically despite a larger number of parameters. Notably, DEFLECT achieved superior performance with 5-10x fewer parameters compared to low-rank-based PEFT methods.
In particular, DEFLECT has empirically demonstrated its efficiency in a wide range of environmental tasks, from pollutant monitoring to the segmentation of areas burnt by forest fires. \vspace{0.1cm}

In future works, we would like to further experiment DEFLECT on hyperspectral satellite images, whose spectral dimension provides richer information on the physical properties of land, water and atmosphere than multispectral data. Moreover, we would like to extend DEFLECT to multi-temporal satellite images and SAR satellite time series. Finally, we believe that DEFLECT could be applied with standard foundation models, for instance to integrate depth information for segmentation tasks.
In conclusion, we see a huge potential for DEFLECT to adapt GFMs to multi-source satellite data and to environmental tasks.

\small
\bibliographystyle{ieee_fullname}
\bibliography{main}

\clearpage
\appendix

\begin{center}
{\Large \bf Supplementary Material \par}
\end{center}

\section{Datasets Details}

\noindent
\textbf{So2Sat} is designed to detect environmental and urban dynamics and land cover changes in cities worldwide using remote sensing data. It combines Sentinel-1 (SAR) and Sentinel-2 (multispectral) imagery to track urban expansion and environmental changes. The dataset contains 17 bands, with imagery captured or upsampled at a 10-meter spatial resolution. The data is divided into 19,992 training images, {1,000} validation images, and 1,000 test images, each patch being 32x32 pixels. We selected, for our purpose, the Sentinel-2 bands. \\

\noindent 
\textbf{Brick Kiln} focuses on detecting and monitoring brick kilns, which are often associated with environmental pollution in rural areas. This dataset uses 13 bands from Sentinel-2 imagery. The images are 64x64 pixels in size, with a 10-meter resolution, and consist of around 15,000 samples for training and around 2,000 for validation and test. \\

\noindent
\textbf{ForestNet} focuses on forest monitoring, aimed at identifying and classifying forested areas to support environmental conservation and management. It utilizes Landsat-8 imagery with 12 bands, including both multispectral and thermal bands, at a 15-meter resolution. The dataset has more than 8,000 patches, each 332x332 pixels. \\

\noindent
\textbf{MADOS} is focused on detecting marine pollution, such as oil spills and debris, in oceanic environments. It contains Sentinel-2 imagery with bands 1-8A, 11, and 12 (8 bands total), covering various sea surface features from 174 scenes. The dataset is segmented into 2803 tiles, each measuring 240x240 pixels. \\

\noindent
\textbf{HLS BurnScars} provides imagery for identifying burn scars, a critical task for monitoring the effects of wildfires on ecosystems and land. It uses Harmonized Landsat and Sentinel-2 (HLS) data with 6 bands: Blue (B02), Green (B03), Red (B04), NIR (B8A), SW1 (B11), and SW2 (B12). The imagery is available at a 30-meter spatial resolution, and the dataset consists of 804 scenes, each with a size of 512x512 pixels. \\

\noindent
Each of these datasets plays a significant role in environmental monitoring using satellite imagery. They are designed to tackle a wide range of issues, including urban expansion, pollution, deforestation, forest management, and the impact of natural disasters like wildfires and oil spills. These datasets are essential for advancing our ability to monitor and manage the Earth's changing environments.

\section{On DEFLECT implementation}

\begin{table}[ht]
\caption{Ablation study about the computation of the spectral patch embeddings on MADOS \label{tab:spectral_patch_embed}}
\centering
\resizebox{\columnwidth}{!}{
\begin{tabular}{l c c}
Model & w/ projection (default) & w/o projection \\ \toprule
Scale-MAE & {50.6} & 50.3 \\
DINO-MC & 51.6 & 48.5 \\
Cross-Scale MAE & {38.2} & 38.1 \\
\end{tabular}
}
\end{table}

\noindent 
\textbf{Reprojecting the features} In the default setting of DEFLECT, the spectral pacth embeddings pass through a linear layer shared across attention blocks, such that the dimension of the spectral embeddings matches the dimension of the spatial embeddings. In some cases (depending on the number of multispectral channels), we can remove the projection layer, and only select the right number of statistics to have the right dimension. Table \ref{tab:spectral_patch_embed} shows the difference of test metrics on MADOS with and without the projection layer. It seems that having a projection layer can slightly improve the performance of DEFLECT. \\

\noindent
\textbf{Pixel-set encoding} In our setting, the pixel-set encoding module randomly samples 10\% of pixels within a patch (as we assumed that the spectral information was redundant within neighboring pixels). Table \ref{tab:pixel-set-encoding} shows the test metrics obtained on MADOS and BurnScars, by sampling 50\% of the pixels, that do not significantly differ from a 10\% sampling. These results confirm our hypothesis that there is strong spectral redundancy within a patch.

\begin{table}[ht]
\caption{Ablation study about the pixel-set encoding module}
\label{tab:pixel-set-encoding}
\centering
\resizebox{1.0\columnwidth}{!}{
\label{tab:pixel-set-encoding}
\begin{tabular}{lcccc}
                & \multicolumn{2}{c}{\begin{tabular}[c]{@{}c@{}}MADOS\\ (mIoU)\end{tabular}} & \multicolumn{2}{c}{\begin{tabular}[c]{@{}c@{}}BurnScars\\ (IoU)\end{tabular}} \\ \midrule 
Model           & 10\%*                                & 50\%                                & 10\%*                                  & 50\%                                 \\ \midrule
Scale-MAE       & 50.6                                 & 51.8                                & 77.3                                   & 78.1                                 \\
DINO-MC         & 51.6                                 & 50.3                                & 75.6                                   & 75.0                                 \\
Cross-Scale MAE & 38.2                                 & 38.5                                & 70.6                                   & 68.9                                
\end{tabular}
}
\end{table}

\begin{figure*}[t]
    \centering
    \begin{subfigure}[t]{0.22\textwidth}
        \centering
        \includegraphics[width=\linewidth]{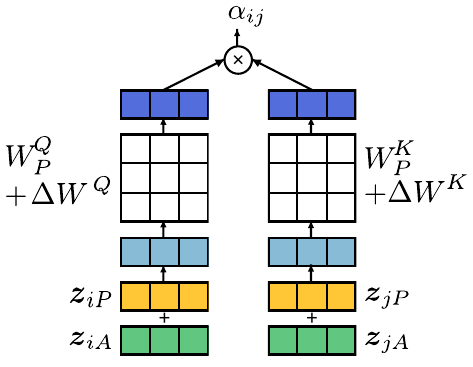}
         \caption{Low-rank adaptation with standard self-attention}
     \end{subfigure}
     \hfill
     \begin{subfigure}[t]{0.77\textwidth}
         \centering
         \includegraphics[width=\linewidth]{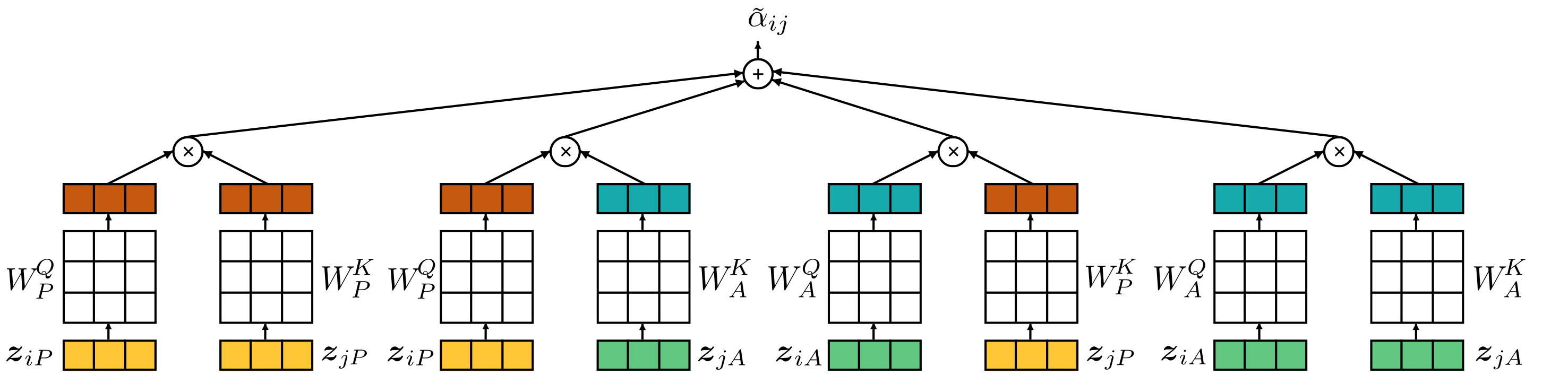}
         \caption{DEFLECT adaptation with untangled attention}
     \end{subfigure}
     \caption{Illustration of standard attention and our untangled attention module in the context of PEFT. \label{fig:att_comp}}
\end{figure*}

\noindent
\textbf{Standard attention VS Untangled Attention} Fig. \ref{fig:att_comp} illustrates the differences between standard attention and our uAtt module.\\

\begin{figure*}
    \centering
    \includegraphics[width=\linewidth]{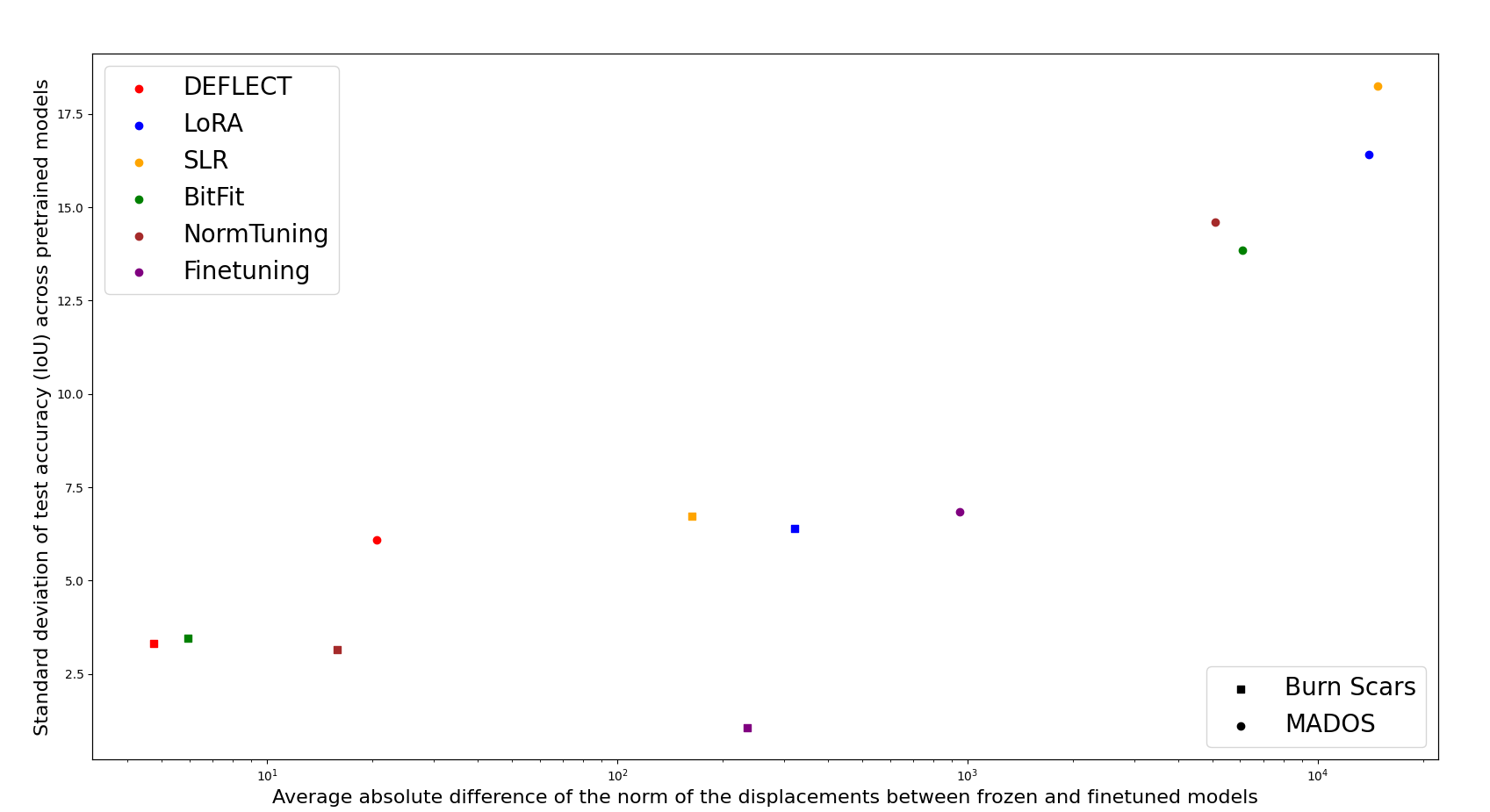}
    \caption{Standard deviation across models as a function of the average absolute difference of norm displacement.}
    \label{fig:enter-label}
\end{figure*}

\noindent
\textbf{Norm of the displacements}
Figure \ref{fig:enter-label} suggests a general trend where larger changes in the norm of displacement between frozen and fine-tuned models lead to increased variability in test accuracy. This indicates that significant modifications to the displacement norm can negatively impact performance consistency across pretrained models. Notably, DEFLECT, which barely changes the norm of the displacement (after the first adapted layer, the norm can change with respect to the frozen pretrained GFM), exhibits lower standard deviation in test accuracy, demonstrating more robust results. This behavior highlights the potential advantage of controlling displacement norm variations and will be further investigated in future work.

\section{Initialization strategies} 

\textbf{HLSBurnScars} To provide insights about the mechanisms of GFM adaptation to multispectral images, we investigated RGB weight initialization strategies, a topic often overlooked. 
The default method, widely used in tools like \textit{timm}\cite{rw2019timm}, involves repeating RGB weights across all bands, regardless of their physical meaning. 
An alternative, introduced by USat \cite{irvin2023usat}, initializes the RGB bands with pretrained weights while assigning random initialization to the others. 
For consistency with related works~\cite{Scheibenreif_2024_CVPR}, we opted for the first method, here called \textit{Repeat}, over the \textit{RGB+random} approach. Tab.~\ref{tab:hls_init} summarizes the results on BurnScars, showing that \textit{Repeat} generally achieved higher scores, though the performance varied significantly across PEFT methods.
For instance, with Scale-MAE, \textit{Repeat} yielded a mIoU of 80.1\% under SLR (w.r.t 78.3\% obtained with \textit{RGB+random}) but dropped to 70.5\% when using DINO-MC (w.r.t 73.2\% with \textit{RGB+random}). 
This variability suggests that while \textit{Repeat} can offer slight advantages, neither approach provides robust, consistent results across methods. 
DEFLECT, that circumvents the choice of an initialization strategy, may thus provide a more reliable framework, avoiding the need for a sensitive selection of hyperparameters. Further results are detailed in the Supplementary Material.

\begin{table}[h!]
    \caption{Performance metrics on BurnScars for two initialization strategies across different PEFT methods, showing similar but inconsistent results. This variability suggests that alternative approaches, such as DEFLECT, which do not depend on specific initialization schemes, may offer more stable and reliable performance in geospatial tasks.}
    \centering
    \resizebox{\columnwidth}{!}{%
    \begin{tabular}{l l c c}
        Model & Tuning Strategy & Repeat & \makecell{RGB+\\random} \\ \toprule
        
        \rowcolor{gray!20} \multirow{4}{*}{Scale-MAE} & Finetuning (Oracle) & 79.1 &  75.5\\
        & Frozen & 76.2 &  68.9\\
        & BitFit & 76.0 &  72.0\\\
        & SLR & 80.1 &  78.3\\
        \hline
        
        \rowcolor{gray!20} \multirow{4}{*}{DINO-MC} & Finetuning (Oracle) & 76.5 &  76.9\\
        & Frozen & 70.4 &  66.9\\
        & BitFit & 69.8 &  66.6\\
        & SLR & 70.5 &  73.2\\
        \hline

        \rowcolor{gray!20} \multirow{4}{*}{Cross-Scale MAE} & Finetuning (Oracle) & 78.1 &  75.8\\
        & Frozen & 68.0 &  59.9\\
        & BitFit & 67.9 &  68.5\\
        & SLR & 63.7 &  68.2\\
        
    \end{tabular}%
    }
    \label{tab:hls_init}
\end{table}

\textbf{MADOS} Table \ref{tab:mados_init} reveals that for MADOS, the "RGB+random" initialization seems to perform slightly better across most methods. This trend further supports our hypothesis that different initialization strategies are not consistently reliable across different tasks. 
For example, "RGB+random" leads to higher mIoU scores in Scale-MAE (e.g. 53.1\% vs. 47.0\% for finetuning) and DINO-MC (e.g. 64.26\% vs. 61.6\% for finetuning). However, when we look at BurnScars (as shown in Table 5 in the main paper), the "Repeat" initialization yields better results on average, further emphasizing the variability of these initialization strategies. \\

\noindent
This inconsistency confirms our key point: relying on specific initialization strategies is not ideal for stable, robust performance. Instead, DEFLECT, which operates independently of initialization schemes, proves to be a more reliable approach across diverse tasks, as it yields more stable results.

\begin{table}[h!]
    \caption{Performance metrics on MADOS for two initialization strategies across different PEFT methods. The table highlights the slight superiority of the RGB+random initialization, though results are inconsistent across datasets. This further supports our argument that reliance on initialization strategies can lead to unstable performance. In contrast, DEFLECT, which does not depend on specific initialization, demonstrates more reliable and stable results.}
    \centering
    \resizebox{\columnwidth}{!}{%
    \begin{tabular}{l l c c}
        Model & Tuning Strategy & Repeat & \makecell{RGB+\\random} \\ \toprule
        
        \rowcolor{gray!20} \multirow{4}{*}{Scale-MAE} & Finetuning (Oracle) &  47.0 &  53.1\\
        & Frozen &  36.0 &  44.4\\
        & BitFit &  19.4 &  48.1\\
        & SLR &  46.5 &  46.9\\
        \hline
        
        \rowcolor{gray!20} \multirow{4}{*}{DINO-MC} & Finetuning (Oracle) &  61.6 & 64.3 \\
        & Frozen &  51.8 &  53.9\\
        & BitFit &  53.0 &  54.1\\
        & SLR &  3.5 &  5.8\\
        \hline

        \rowcolor{gray!20} \multirow{4}{*}{Cross-Scale MAE} & Finetuning (Oracle) &  47.2 & 50.1 \\
        & Frozen &  41.4 &  43.5\\
        & BitFit &  40.3 &  39.8\\
        & SLR &  35.6 &  36.1\\
        
    \end{tabular}%
    }
    \label{tab:mados_init}
\end{table}

\end{document}